\documentclass[10pt,twocolumn,letterpaper]{article}
\pdfoutput=1
\usepackage{cvpr}
\usepackage{times}
\usepackage{epsfig}
\usepackage{graphicx}
\usepackage{amsmath}
\usepackage{amssymb}
\usepackage{epstopdf}
\usepackage{multirow}
\usepackage{algorithm}
\usepackage{algorithmic}
\usepackage{authblk}
\usepackage{MnSymbol,wasysym}

\usepackage[export]{adjustbox}

\newtheorem{theorem}{Theorem}[section]

\newtheorem{definition}[theorem]{Definition}

\newtheorem{result}[theorem]{Result}

\newtheorem{rules}[theorem]{Rule}

\newcommand{\norm}[1]{\left\lVert#1\right\rVert}
\newcommand{\qed}{\nobreak \ifvmode \relax \else
      \ifdim\lastskip<1.5em \hskip-\lastskip
      \hskip1.5em plus0em minus0.5em \fi \nobreak
      \vrule height0.75em width0.5em depth0.25em\fi}

\usepackage[pagebackref=true,colorlinks,bookmarks=false]{hyperref}

\cvprfinalcopy 

\begin{document}

\title{Mapping, Localization and Path Planning for\\Image-based Navigation using Visual Features and Map}
\author[1]{Janine Thoma}
\author[1]{Danda Pani Paudel}
\author[1]{Ajad Chhatkuli}
\author[1]{Thomas Probst}
\author[1,2]{Luc Van Gool}
\affil[1]{Computer Vision Laboratory, ETH Zurich, Switzerland}
\affil[2]{VISICS, ESAT/PSI, KU Leuven, Belgium}

\maketitle
\thispagestyle{empty}
\begin{abstract}
Building on progress in feature representations for image retrieval, image-based localization has seen a surge of research interest.
Image-based localization has the advantage of being inexpensive and efficient, often avoiding the use of 3D metric maps altogether.
That said, the need to maintain a large number of reference images as an effective support of localization in a scene, nonetheless calls for them to be organized in a map structure of some kind.

The problem of localization often arises as part of a navigation process. We are, therefore, interested in summarizing the reference images as a set of landmarks, which meet the requirements for image-based navigation. A contribution of this paper is to formulate such a set of requirements for the two sub-tasks involved: map construction and self-localization.
These requirements are then exploited for compact map representation and accurate self-localization, using the framework of a network flow problem. During this process, we formulate the map construction and self-localization problems as convex quadratic and second-order cone programs, respectively.
We evaluate our methods on publicly available indoor and outdoor datasets, where they outperform existing methods significantly\footnote{Code: \scriptsize{\url{https://github.com/janinethoma/}}}.
\end{abstract}

\section{Introduction}
Vision-based navigation is one of the key components of robotics, self-driving cars and many mobile applications. It is tackled either by using a 3D map representation such as in Structure-from-Motion (SfM) based methods~\cite{Irschara2009,Li2010location,Sattler2017,Sattler2017pami,Zeisl2015,choudhary2012visibility} and Simultaneous Localization and Mapping (SLAM) methods~\cite{Mouragnon2006,davison2007monoslam,castle2008video,Bresson2017,eade2006scalable} or  by using a map purely represented with geo-tagged images~\cite{arandjelovic2016netvlad,Simonyan2014,Arandjelovic2014,Kim2017}. In contrast to SfM and SLAM-based methods, localization by image retrieval (or simply image-based localization) is inexpensive, with a simple map representation, which also scales better in larger spaces~\cite{arandjelovic2016netvlad,Sattler2017}. The problem of image-based localization is posed as the matching of one or more query images taken at unknown locations to a set of reference images captured at known locations in a map. Recent developments in learning image feature representations for object and place recognition~\cite{krizhevsky2012imagenet,Simonyan2014,Arandjelovic2014,arandjelovic2016netvlad} have made image retrieval a viable method for localization. Despite the increased interest, image-based navigation methods are largely error-prone due to matching inaccuracies~\cite{Anoosheh2018}. Some existing methods address this by learning better feature representations for place recognition~\cite{arandjelovic2016netvlad,Simonyan2014,Arandjelovic2014,Kim2017}. Nonetheless, errors in matches cannot be avoided in realistic settings with changes in illumination, camera pose and dynamic objects~\cite{Anoosheh2018}. Methods that directly regress poses~\cite{Taira2018,Kendall2015,Kendall2017} naturally run into similar problems. We argue that, in addition to feature representation, the success of localization in navigation is determined by several other key factors. In particular, current methods do not adequately address the problem of map representation. 

Many methods use a large (or even complete) reference image set in order to localize a given query image~\cite{Arandjelovic2014,Kim2017,Anoosheh2018}. 
Although a large reference set has a higher chance of a similar (in pose and illumination) reference and query image pair to exist, it not only leads to higher memory requirements but may also become sub-optimal for the matching process. Another important neglected aspect in image based localization is the order of query image sequences, which is the key to the success of visual SLAM methods. Unlike SLAM, localization by retrieval often works with a much sparser sequence of query images. Exploiting information from such interleaved image sequences is very challenging. In this context, \cite{milford2012seqslam} localizes a sequence of query images by assuming a linear change of features over time. However, this assumption is rather naive, since it fails as soon as some objects appear (or disappear) in images. As a consequence, we are interested in answering the question of what are the desired criteria of a good map representation for image retrieval-based navigation? And how can we benefit from such a representation during image localization?

In this paper, we address the task of navigation on a map where there exist geometric relationships between images or landmarks. Given visual features of images and image locations of the reference set, we identify three key problems: map construction by image selection, path planning, and localization using a history of image matches for multiple images. In particular, we provide new methods for map construction and matching multiple images to the reference images of the map. We present the construction and representation of the map as image landmark selection from a sequence of images using the principles of optimal transport. For that purpose we introduce rules that direct how images should be selected for the map representation and derive the costs accordingly. We model the rules as a problem of computing flow from source images to target images given the image geometric locations and the visual features and solve it using Quadratic Programming (QP). Our second contribution is about the localization of multiple query images on the map, where we model the problem as bipartite graph matching. We solve the localization by computing a flow between the landmark images as the sources and the query images as the targets in the bipartite graph, using Second Order Cone Programming (SOCP). We evaluate both landmark image selection and localization on publicly available indoor and outdoor datasets, and show that we significantly outperform the state-of-the-art.

\section{Related Work}
We briefly describe some relevant works on map-based localization by image retrieval. We do not discuss the alternative approach of navigation based on a pre-built 3D map using SfM or SLAM~\cite{Irschara2009,Li2010location,Sattler2017,choudhary2012visibility,Sattler2017pami,Zeisl2015}. Although both map-building and navigation play important roles in localization by image retrieval, most research interests have been directed towards learning better features~\cite{arandjelovic2016netvlad,Simonyan2014,Arandjelovic2014,Kim2017}. Nonetheless, some progress has been made on modeling the map and matching. In particular, \cite{kosecka2004vision} perform map-building by uniformly sampling video streams for images and improve matching by interpreting the map as a Hidden Markov Model. However, they do not model the temporal relation of input images in the matching process. Moreover the uniform sampling for map-building may not be the optimal approach. \cite{vysotska2016lazy} model the query image sequence to map sequence matching as a directed graph problem where the temporal continuity is exploited. A similar strategy is also pursued by \cite{milford2012seqslam}. However, previous works do not fully consider the context of the navigation process when tackling the problems of map-building, localization and path planning. In the following sections, we define the preliminaries for modeling these problems using the theory of optimal transport and solve the problems based on the rules we develop.

\section{Preliminaries}
Let us consider a graph $\mathcal{G} = (\mathcal{V},\mathcal{E})$ 
with a set of vertices  $\mathcal{V}=\{v_i\}$  and a set of directed edges $\mathcal{E}=\{e_{ij}\}_{i\not=j}$. 
For the edge $e_{ij}\in\mathcal{E}$, we define the flow capacity and the flow cost rate $u_{ij},c_{ij}\in\mathbb {R}^{+}$, respectively. Let $y_{ij}\in [0,u_{ij}]$ be the flow for $e_{ij}\in\mathcal{E}$,  such that the flow of an edge is non-negative and cannot exceed its capacity.  For each vertex $v_i\in \mathcal{V}$, we define the total outgoing flow $\underline{y}_i=\sum_{e_{ij}}y_{ij}$  and the total incoming flow $\overline{y}_i=\sum_{e_{ji}}y_{ji}$, such that the net flow is $y_i=\underline{y}_i-\overline{y}_i$ and the absolute flow is $\hat{y}_i=\underline{y}_i+\overline{y}_i$. We consider two sets $\mathcal{S}, \mathcal{T}\subset\mathcal{V}$  for source and target vertices respectively, such that   $\mathcal{S}\cap\mathcal{T}=\emptyset$.
For each source vertex $v_i\in \mathcal{S}$, we are given the  
net outgoing flow  $s_i\in\mathbb{R}^+$. Similarly, $t_i\in\mathbb{R}^+$ is the given net incoming flow of target vertex $v_i\in \mathcal{T}$. For the remaining vertices, we  apply the rule of conservation of flows: the sum of the flows entering a vertex must equal the sum of the flows exiting a vertex. We also ensure that the flow between the sources and targets are conserved by imposing the flow constraint $\sum_{v_i\in\mathcal{S}}s_i=\sum_{v_i\in\mathcal{T}}t_i$. 
Now, we wish to transport the source flows $\{s_i\}$ to the target flows $\{t_i\}$, with minimal transportation cost, by solving the following optimization problem.  
\begin{equation}
\begin{aligned}
& \underset{y_{ij}}{\textnormal{min}}
& &  \sum_{e_{ij}\in\mathcal{E}} c_{ij}y_{ij}, \\
& \textnormal{~s.t.}  & & 0\leq {y}_{ij}\leq u_{ij},\,\,\,\,\,\,\,\,\,\,\,\,\,\,\,\,\,\,\,\,\,\,\,\,\,\,\,\,\,\, \forall e_{ij}\in \mathcal{E}\\
& & & y_{i}= 
\begin{cases}
    s_i & \forall v_i\in\mathcal{S}\\
    -t_i & \forall v_i\in\mathcal{T}\\
    0 & \forall v_i\in \mathcal{V}\setminus (\mathcal{S}\cup\mathcal{T})
\end{cases}
\end{aligned}
\label{eq:flowFormulaiton_0}
\end{equation}
The problem of~\eqref{eq:flowFormulaiton_0} is convex and can be solved using Linear Programming (LP). There exist various off-the-self solvers~\cite{Mosek,SeDuMi} that offer very efficient LP solutions.

\section{ Image-based Navigation}
We rely only on images and the scene topology for all three sub-tasks of navigation---map representation, path planning, and self localization.
During these processes, visual features of images and their locations on the topological map are considered. In the following, we provide the exact problem setup addressed in this paper, followed by our solutions for each of the three sub-tasks.

\subsection{Problem Setup}\label{subSec:problemSetup}
We consider a map $\mathcal{M}\subset \mathbb{R}^2$ and a set of images $\mathcal{I} = \{\mathcal{I}_i\}_{i=1}^n$ with their location coordinates $\mathcal{X} = \{\mathsf{x}_i\in \mathcal{M}\}_{i=1}^n$ and visual features $\mathcal{F} = \{\mathsf{f}_i\}_{i=1}^n$. Using this information, we construct  a graph $\mathcal{G} = (\mathcal{V},\mathcal{E})$ where the set of vertices  $\mathcal{V}=\{v_i\}$ represent images $\mathcal{I}_i, i = 1,\ldots,n$ and the set of directed edges $\mathcal{E}=\{e_{ij}\}_{i\not=j}$ represent pairwise relations between images $\mathcal{I}_i$ and $\mathcal{I}_j$. Efficient navigation demands a compact representation of $\mathcal{G}$, supporting path planning and the self-localization of image sequences.

\subsection{Map Representation}
\label{subsec:MapRepresentation}
For a given set of vertices $\mathcal{V}=\{v_i\}_{i=1}^n$, we wish to summarize them as a set of landmarks $\mathcal{V}'=\{v'_i\}_{i=1}^m$ such that $\mathcal{V}'\subset \mathcal{V}$.  To do so, we first define the following measure,
\begin{equation}\label{eq:kx}
k_x= \begin{aligned}
&\underset{i=1,\ldots,m}{\textnormal{argmin}}
&\mathbf{d}(\mathsf{x},v_i'),
\end{aligned}
\end{equation}
where $\mathbf{d}(\mathsf{x},v_i)$ is the distance measure between $\mathsf{x}$ and $\mathsf{x}_i$ of the vertex $v_i$. 
Here, $k_x$ is the index of the vertex in $\mathcal{V'}$ which is geometrically closest to the point $\mathsf{x}$. While summarizing the landmarks, we consider the following four rules.
\begin{rules}[Geometric Representation]\label{rule:geometric}
Landmarks must be well distributed geometrically, i.e.\ the selected landmarks must minimize the following, 
\begin{equation}\label{eq:geometric}
 \begin{aligned}
\underset{\mathcal{V}'}{\textnormal{min}}&&\underset{\mathsf{x}\in \mathcal{M}}{\textnormal{max}} &
&\mathbf{d}(\mathsf{x},v_{k_x}').
\end{aligned}
\end{equation}
\end{rules}
\begin{rules}[Visual Representation]\label{rule:visual}
 Landmarks must be useful for localizing images using their visual features. More precisely, all images must have a small feature distance to the geometrically closest landmark, i.e.\ for the feature distance $\mathbf{d}(\mathsf{f}_i, \mathsf{f}_j)$, landmarks must also respect,
\begin{equation}\label{eq:visual}
 \begin{aligned}
\underset{\mathcal{V}'}{\textnormal{min}}&&\underset{ \{
\mathsf{x,f} \}\in \mathcal{V}}{\textnormal{max}} &
&\mathbf{d}(\mathsf{f},\mathsf{f}_{k_x}').
\end{aligned}
\end{equation}
\end{rules}

\begin{rules}[Navigation Assurance]\label{rule:navigation}
Landmarks must support navigation from any source to to any target location, using only visual features. In other words, the next landmark along the path must not only be close, it must also be distinct from the current one, to avoid confusion, i.e.\ if $\mathcal{P}=\{v'_l\}_{l=1}^q \subset \mathcal{V}'$ is the ordered sequence of landmarks along a path, two consecutive landmarks must be within the distance $\alpha$ such that, 
\begin{equation}\label{eq:navigation}
 \begin{aligned}
 \mathbf{d}(\mathsf{x}'_l,\mathsf{x}'_{l+1})\leq \alpha,  && \forall 
\mathsf{x}'_l\in \mathcal{P},
\end{aligned}
\end{equation}
and their visual features must be distinct such that, 
\begin{equation}\label{eq:R3NavFearure}
 \begin{aligned}
\underset{\mathcal{V}'}{\textnormal{max}}&&\underset{
\mathsf{f}'_l, \mathsf{f}'_{l+1}\in \mathcal{P}'}{\textnormal{min}} &
&\mathbf{d}(\mathsf{f}'_l,\mathsf{f}'_{l+1}).
\end{aligned}
\end{equation}
This  ensures that the navigation process can find the next landmark  without getting confused  with the previous one. 
\end{rules}
\begin{rules}[Map Compactness]\label{rule:compctness} 
The number of landmarks must be small, i.e.\ $|\mathcal{V'}| \leq N$, for maximally $N$ landmarks.
\end{rules}
Landmark summarization for image-based navigation is a multi-objective problem which favours the above four rules. 

\subsection{Path Planning}
The task of path planning is to choose an ordered set of landmarks that help to travel from a given source to a target location along the shortest path using only the landmark images.
Since the rules of map representation already ensure a good set of landmarks, the task of path planning simply becomes a problem of finding the shortest path along the selected landmarks. Such a path can be found using existing methods such as Dijkstra's algorithm.

\subsection{Self Localization}
Given a sequence of images and landmarks along a path, the task of self-localization is equivalent to finding the most consistent match. We assume that an ordered sequence of images $\mathcal{I}_p=\{\mathcal{I}_l\}_{l=1}^q$, captured along a path are given. We wish to localize these images by matching them to landmarks $\mathcal{V}'$.
We formulate self-localization as a graph matching problem between $\mathcal{P}=\{p_l\}_{l=1}^q$ representing $\mathcal{I}_p$ and $\mathcal{V}'=\{v_i'\}_{i=1}^m$.
Let $\mathcal{B}:l\rightarrow l'$ be a map that generates the desired matching pairs $\{p_l,v'_{l'}\}_{l=1}^q$ of sequence images and landmark images. For the purpose of self-localization, we want the matching process to favor the following two rules.  

\begin{rules}[Visual Matching]\label{rule:visulaMatching}
The visual distance between matched pairs must be minimized, i.e.\ if $\{\mathsf{f}_l,\mathsf{f'}_{l'}\}$ are the visual features coming from the pair $\{p_l,v'_{l'}\}$, the mapping corresponding to the best matching is found as
\begin{equation}\label{eq:visualMatch}
 \begin{aligned}
\underset{\mathcal{B}}{\textnormal{min}}&
&\sum_l{\mathbf{d}(\mathsf{f}_l,\mathsf{f}'_{l'})}.
\end{aligned}
\end{equation}
\end{rules}
\begin{rules}[Geometric Matching]\label{rule:geometricMatching}
Neighbours of $p_l$ must be matched to  the neighbours
of $v'_{l'}$ or $v'_{l'}\in\mathcal{V}'$ itself. i.e.\ 
\begin{equation}\label{eq:geometricMatch}
 \begin{aligned}
v'_{l'-1},v'_{l'+1} \in\mathcal{N}(v'_{l'})\cup v'_{l'},&& \forall p_l\in\mathcal{P}.
\end{aligned}
\end{equation}
\end{rules}

\section{ Map Construction using Network Flow}
\label{sec:MapConstruction}
\begin{figure}
    \includegraphics[width=\linewidth]{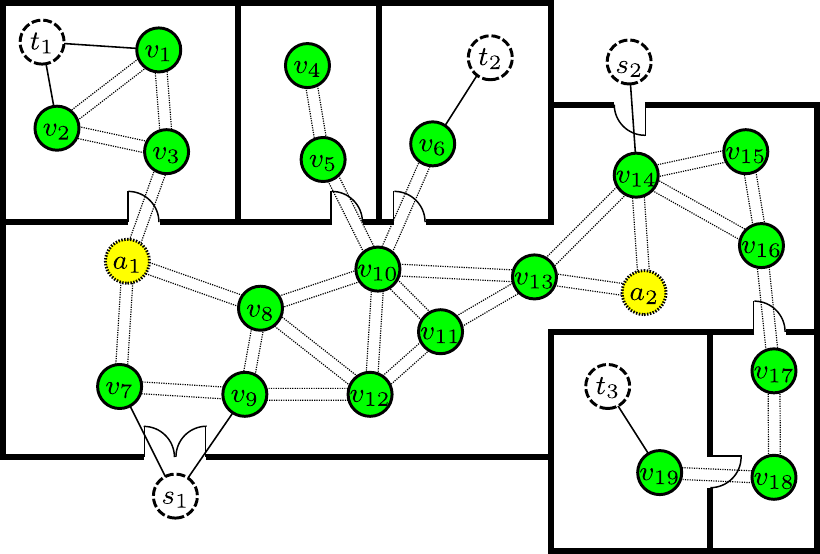}
    \caption{Visualization of Graph $\mathcal{G}=\{\mathcal{V,E}\}$ for map construction with sources $s_i \in \mathcal{S}$, targets $t_i \in \mathcal{T}$, anchor points $a_i \in \mathcal{A}$, and remaining image vertices $v_i \in \mathcal{V}$.}
    \label{fig:mapping}
\end{figure}

To represent the map using only images, 
we define the graph $\mathcal{G}=\{\mathcal{V,E}\}$, as discussed in the previous section and as visualized in Fig.~\ref{fig:mapping}.  
Any edge $e_{ij} \in \mathcal{E}$ represents the relationship between $\mathcal{I}_i$ and $\mathcal{I}_j$, using the flow  capacity $u_{ij}$  and cost rate $c_{ij}$  defined as,
\begin{equation}
    u_{ij}=\lambda_x\mathbf{d}(\mathsf{x}_i, \mathsf{x}_j) \text{\,\,\,\,\,and\,\,\,\,\,} c_{ij}=\lambda_f/\mathbf{d}(\mathsf{f}_i, \mathsf{f}_j),\label{eq:capCostDef}
\end{equation}
for weights $\lambda_x$ and $\lambda_f$ associated to geometric and visual measures. Recall that  $\mathbf{d}(\mathsf{x}_i, \mathsf{x}_j)$ and  $\mathbf{d}(\mathsf{f}_i, \mathsf{f}_j)$  respectively are the geometric and visual distances between images $\mathcal{I}_i$ and $\mathcal{I}_j$. Here,  we first define the landmark selection process for map representation, in the context of network flow.
\begin{definition}[Landmarks]
Graph vertices with absolute flow greater than
a given flow threshold $\tau$ are the desired landmarks. i.e.\ the landmarks are, $\mathcal{V}'=\{v_i: \hat{y}_i\geq \tau \}$.
\end{definition}
In the following, we make use of~\eqref{eq:capCostDef} within the formulation of~\eqref{eq:flowFormulaiton_0}, with additional constraints,  in order to obtain landmarks that favour rules~\ref{rule:geometric}--\ref{rule:compctness}. We also provide the reason behind our choice of cost rate and capacity expressed in~\eqref{eq:capCostDef}. 

\subsection{Geometric Representation}
\label{subseq:Anchors}
The geometric representation Rule~\ref{rule:geometric} is indeed the well known $k$-Center Problem, which in itself is NP-hard. However, there exist simple greedy approximation algorithms of $\mathcal{O}(n)$ complexity that solve the $k$-Center Problem with an approximation factor of 2. We use a similar approach to choose a set of anchor points $\mathcal{A}\subset \mathcal{M}$ by solving,
\begin{equation}\label{eq:k-centerProblem}
\begin{aligned}
\underset{\mathcal{A}}{\textnormal{min}}&&|\mathcal{A}|&&\text{s.t.}&& \mathbf{d}(\mathsf{x},\mathcal{A})\leq r/2, \forall \mathsf{x}\in\mathcal{M}
\end{aligned}
\end{equation}
for radius $r$ and point-to-set distance $\mathbf{d}(\mathsf{x},\mathcal{A})$. Note that the distance is constrained by $r/2$ to compensate the approximation factor of 2. Using the obtained set of anchor points, we impose the following constraint on the absolute flow to favour the geometric representation Rule~\ref{rule:geometric},
\begin{equation}\label{eq:visualConstarint}
 \begin{aligned}
    \sum_{v_i\in\mathcal{N}(a)}{\hat{y}_i}\geq t_g,&&&\forall a\in\mathcal{A}.
 \end{aligned}
\end{equation}
for neighbourhood flow threshold $t_g$ and neighbouring vertices $\mathcal{N}(a)\subset \mathcal{V}$ of the anchor point $a$ within a radius $r$. The constraint in~\eqref{eq:visualConstarint} ensures flow around every anchor point, thus encouraging the landmarks to be well distributed. In fact, one can alternatively maximize $t_g$ to guarantee the feasibility of the network flow problem, by adding a term $-\lambda_gt_g$ to the original cost  of~\eqref{eq:flowFormulaiton_0} for a constant weight $\lambda_g$. 

\subsection{Visual Representation}
\label{subseq:Sensitivity}
The rule of visual representation demands no image to be visually too far from its geometrically closest landmark. Therefore, all nodes with distinct visual features in a local neighbourhood must have significant absolute flow. We ensure such flow by introducing the flow sensitivity $\rho_{ij}$ for every edge $e_{ij}\in\mathcal{E}$.   
The flow sensitivity controls the cost rate as flow approaches capacity, such that the new cost rate is given by,
\begin{equation}\label{eq:newRate}
    b_{ij}= c_{ij}+\frac{y_{ij}\rho_{ij}}{1-y_{ij}/u_{ij}},
\end{equation}
for base cost rate $c_{ij}$ and sensitivity $\rho_{ij}$. We define the sensitivity using the feature distribution around a vertex as follows,
\begin{equation}\label{eq:senDef}
    \rho_{ij}= 1 - \frac{ \mathbf{d}(\mathsf{f}_i,\mathsf{f}_j)}{\sum_{k\in\mathcal{N}(\mathsf{x}_i)}{\mathbf{d}(\mathsf{f}_i,\mathsf{f}_k)}}.
\end{equation}
The sensitivity encourages the flow to spread before the maximum capacity of the cheapest edge is used. This is particularly important when a diverse set of visual features is clustered together geometrically. In such cases, the risk is that the flow primarily passes though only one vertex thus selecting only one landmark, since both  incoming and outgoing edges offer low cost and sufficient capacity.  This violates the visual representation rule. In such circumstances, sensitivity encourages the flow to spread around such that more than one landmark is selected favouring Rule~\ref{rule:visual}. 
Note that the sensitivity is high for higher feature diversity.
On the other hand, if there is only one distinct visual feature in a neighbourhood, the flow sensitivity of the edge to that vertex is very low. Using~\eqref{eq:newRate} and~\eqref{eq:senDef},  the new cost corresponding to any edge $e_{ij}\in\mathcal{E}$ can be expressed as,
\begin{equation} 
\begin{aligned}
     y_{ij}b_{ij} = y_{ij}c_{ij}+z_{ij},&&\text{with} &&
    z_{ij}\geq \frac{y_{ij}^2\rho_{ij}}{1-y_{ij}/u_{ij}},
\end{aligned}
\end{equation}
where the inequality  is a rotated cone constraint,
\begin{equation}
    \big(u_{ij}-y_{ij}\big)\big(z_{ij}/(\rho_{ij}u_{ij})\big)\geq y_{ij}^2.
\end{equation}

\subsection{Navigation Assurance}
The formulation of network flow ensures that all flow must transfer from source to sink vertices. Therefore, the network flow problem is already tuned for the navigation task. While encouraging the flow to make bigger geometric jumps, by keeping the capacity directly proportional to the geometric distance (using \eqref{eq:capCostDef}), we ensure that all jumps are smaller than the navigation radius $\alpha$ by constructing the graph such that,
\begin{equation}\label{eq:distLimit}
\begin{aligned}
    \mathbf{d}(\mathsf{x}_i,\mathsf{x}_j)\leq \alpha, && \forall e_{ij}\in\mathcal{E}. 
\end{aligned}
\end{equation}
Furthermore, we minimize the flow between two vertices with similar features, by keeping the cost rate inversely proportional to the  feature distance (using \eqref{eq:capCostDef}). 
This encourages the selection of distinct consecutive features  along the flow path, thus favouring the objective of~\eqref{eq:R3NavFearure}. The construction of a locally connected graph and our choice of cost rate and capacity support the navigation assurance Rule~\ref{rule:navigation}.

\subsection{Map Compactness}\label{subSec:mapCompactness}
Given a threshold $\tau$ on the absolute flow, for a vertex to qualify as a landmark,
we determine a set of landmarks ${\mathcal{V}'=\{v_i: \hat{y}_i\geq \tau \}}$
by controlling source and target flows $\{s_i\}$ and $\{t_i\}$.
Starting from the input/output flow $Y_\mathcal{G} = \sum_\mathcal{s_i\in S}{s_i}=\sum_\mathcal{t_i\in T}{t_i}$,  we gradually increase $Y_\mathcal{G}$ to generate
new landmarks as long as the flow problem remains feasible and $|\mathcal{V'}| \leq N$, for a given upper bound on the number of landmarks $N$.  In this process, the most important landmarks are generated in the beginning. Therefore, one can further control the compactness by choosing the desired number of initial landmarks. 

\subsection{Map Construction Algorithm}
In the following, we present the flow formulation which builds the core of our landmark selection method and summarize our graph representation to map construction process in the form of a landmark selection algorithm. 

\begin{result}
Given a graph $\mathcal{G}=\{\mathcal{V},\mathcal{E}\}$ with cost rate, capacity, and sensitivity $\{c_{ij},u_{ij}, \rho_{ij}\}$ for each edge $e_{ij}\in\mathcal{E}$, a set of anchor points $\mathcal{A}$, source and target vertices $\mathcal{S}$ and $\mathcal{T}$, and a neighbourhood flow threshold $t_g$,  
the flows $\{y_{ij}\}$ required for map reconstruction can be obtained by solving the following network flow problem.
\begin{equation}
\begin{aligned}
& \underset{y_{ij},z_{ij}}{\textnormal{min}}
& &  \sum_{e_{ij}\in\mathcal{E}} (c_{ij}y_{ij}+z_{ij}), \\
& \textnormal{~s.t.}&&\big(u_{ij}-y_{ij}\big)\big(z_{ij}/(\rho_{ij}u_{ij})\big)\geq y_{ij}^2,\,\,\,\,\,\,\,\,\forall e_{ij}\in\mathcal{E},\\
&&& 0\leq {y}_{ij}\leq u_{ij},\,\,\,\,\,\,\,\,\,\,\,\,\,\,\,\,\,\,\,\,\,\,\,\,\,\,\,\,\,\,\,\,\,\,\,\,\,\,\,\,\,\,\,\,\,\,\,\,\,\,\,\,\,\,\,\,\,\,\,\forall e_{ij}\in\mathcal{E},\\
&&&\sum_{v_i\in\mathcal{N}(a)}{\hat{y}_i}\geq t_g,\,\,\,\,\,\,\,\,\,\,\,\,\,\,\,\,\,\,\,\,\,\,\,\,\,\,\,\,\,\,\,\,\,\,\,\,\,\,\,\,\,\,\,\,\,\,\,\,\,\,\,\,\,\forall a\in\mathcal{A},\\
& & & y_{i}= 
\begin{cases}
    s_i, & \forall v_i\in\mathcal{S},\\
    -t_i, & \forall v_i\in\mathcal{T},\\
    0, & \forall v_i\in \mathcal{V}\setminus (\mathcal{S}\cup\mathcal{T}).
\end{cases}
\end{aligned}
\label{eq:flowFormulaiton_final}
\end{equation}
\end{result}
The flow problem of~\eqref{eq:flowFormulaiton_final} is convex and can be solved optimally using Quadratic Programming (QP). In Algorithm~\ref{alg:selectLandmarks}, we summarize the complete process of obtaining landmarks, starting from images with features and locations. Note that the flow problem needs to be solved multiple times to obtain the desired compactness, as discussed in Section~\ref{subSec:mapCompactness}. This can be done either by gradually increasing the input/output flow $Y_\mathcal{G}$ (as discussed earlier), or by performing a bisection search on the parameter $Y_\mathcal{G}$.

\begin{algorithm}
{\small
\caption{\small $\mathcal{V}'$ = selectLandmarks($\mathcal{I}$, $\mathcal{F}$, $\mathcal{X}$, $\mathcal{M}$,  $\mathcal{S}$, $\mathcal{T}$)}
\label{alg:selectLandmarks}
\begin{algorithmic}
 \STATE 1. Construct $\mathcal{G}= \{\mathcal{V},\mathcal{E}\}$ using $\mathcal{I}$, $\mathcal{F}$ and $\mathcal{X}$ (ref. Sec~\ref{subSec:problemSetup}/\eqref{eq:distLimit}).
 \STATE 2. Compute capacity $u_{ij}$ and rate $c_{ij}$ for all $e_{ij}\in\mathcal{E}$, using~\eqref{eq:capCostDef}.
 \STATE 3. Select anchor points $\mathcal{A}\subset \mathcal{M}$, by solving~\eqref{eq:k-centerProblem} for k-centers. 
 \STATE 4. Compute the flow sensitivity  $\rho_{ij}$ for all $e_{ij}\in\mathcal{E}$, using~\eqref{eq:senDef}.
 \STATE 5. Solve the flow problem~\eqref{eq:flowFormulaiton_final} for sources $\mathcal{S}$ and targets $\mathcal{T}$.
 \STATE 6. Derive $\hat{y}_i$ for all $v_i\in\mathcal{V}$ from $y_{ij}$. Return, ${\mathcal{V}'=\{v_i: \hat{y}_i\geq \tau \}}$.
\end{algorithmic}
}
\end{algorithm}

\section{Network Flow for Self Localization}\label{sec:selfLocalization}
\begin{figure}
    \includegraphics[width=\linewidth]{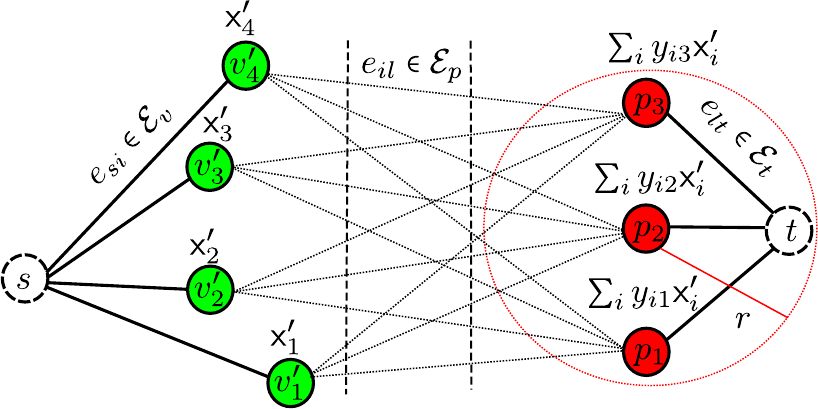}
    \caption{Bipartite graph $\mathcal{G}_b$ with source $s$ and target $t$.}
    \label{fig:selfLocalizationMatching}
\end{figure}

We formulate the self localization of an ordered sequence of images $\mathcal{P}$ with respect to landmarks $\mathcal{V}'$ as a bipartite graph matching problem. For this task, we construct a complete bipartite graph $\mathcal{G}_b=\{\mathcal{V}'$, $\mathcal{P},\mathcal{E}_p\}$ with directed edges $e_{il}\in\mathcal{E}_p$ from $v'_i\in\mathcal{V}'$ to $p_l\in\mathcal{P}$. In addition, we introduce auxiliary source and target vertices $s$ and $t$, respectively.  The source $s$ is connected to  all the vertices $v'_i\in\mathcal{V}$ with directed edges $e_{si}\in\mathcal{E}_v$. Similarly, directed edges $e_{lt}\in\mathcal{E}_t$ connect $p_l\in\mathcal{P}$ to $t$. Using $\mathcal{G}_b$, $s$ and $t$, we represent the flow network using the graph $\mathcal{G}=\{\mathcal{V},\mathcal{E}\}$ with $\mathcal{V}=s\cup\mathcal{V}'\cup \mathcal{P} \cup t$ and $\mathcal{E}=\mathcal{E}_v\cup\mathcal{E}_p\cup\mathcal{E}_t$, as shown in Fig.~\ref{fig:selfLocalizationMatching}. In this section, we solve the bipartite graph matching problem using the network flow formulation of ~\eqref{eq:flowFormulaiton_0},  with  additional  constraints to obtain  matches that respect  Rules~\ref{rule:visulaMatching} and~\ref{rule:geometricMatching}.   

\subsection{Visual Matching}
To obtain visually similar matches, we define the flow cost rate between any landmark and a query image using the visual distance between them. On the other hand, no cost is added for the flow from source to landmarks and from query images to target. Furthermore, we introduce a robust loss for feature matching such that the cost rate  is defined as,   
\begin{equation}
    c_{ij} = h\left(\bf{d}(\mathsf{f}_i,\mathsf{f}_j)\right), \forall e_{ij}\in\mathcal{E}_p;    c_{ij} = 0,\forall e_{ij}\in\mathcal{E}\setminus\mathcal{E}_p, 
    \label{eq:costVisualMatch}
\end{equation}
where $h(.)$ is the Huber loss function. To ensure that an image cannot be matched to more than one landmark, we limit the maximum absolute flow at every query image to one. This translates to the the following capacity constraints,  
\begin{equation}
    u_{ij} = q,\forall e_{ij}\in\mathcal{E}_v; 
    u_{ij} = 1,\forall e_{ij}\in\mathcal{E}\setminus\mathcal{E}_v.
    \label{eq:capacityVisualMatch}
\end{equation}
We allow many query images to be matched to one landmark, by setting the source to landmark capacity higher than one. Additionally,~\eqref{eq:capacityVisualMatch} also ensures the matching of every query image.
\subsection{Geometric Matching}
Recall that we are given only the visual features of the query images, along with the visual features and geometric locations of the landmarks. In this regard, our task is to infer the geometric location of the query images. To do so, for a given flow between the landmarks and the query images, we first define the location of the query images as follows,
\begin{equation}
\mathsf{x}_l=\sum_{v_i\in\mathcal{V}'}{\mathsf{x}_i y_{il}} \text{ for } p_l\in\mathcal{P}.
\label{eq:flowSumGeomMatch}
\end{equation}
Note that the absolute flow of every query image is ${\sum_{i}y_{il}=1}$. 
Therefore,~\eqref{eq:flowSumGeomMatch} guarantees that the query image lies within the convex polytope defined by landmark locations. Now, the geometric matching rule of~\ref{rule:geometricMatching} for navigation radius $r$ and sequential query image pair $\{p_l,p_{l+1}\}$ can be expressed as the following quadratic constraint,
\begin{equation}
    \norm{\sum_{v_{i}\in\mathcal{V}'}{\mathsf{x}_iy_{i(l+1)}} - \sum_{v_{i}\in \mathcal{V}'}{\mathsf{x}_i}y_{il}}\leq r, \forall p_l,p_{l+1}\in\mathcal{P}.
\end{equation}
\subsection{Self Localization Algorithm}
We perform self localization by performing bipartite graph matching, using network flow. 
In the following, we first present the proposed network flow formulation for self localization, as one of our results. Subsequently, we summarize our self localization method as an algorithm. 

\begin{result} 
Consider a  graph $\mathcal{G}=\{\mathcal{V},\mathcal{E}\}$ constructed  using vertices $\mathcal{P}=\{p_l\}_{l=1}^q$ (representing a sequence of images $\mathcal{I}_p$) and landmarks $\mathcal{V}'=\{v_i'\}_{i=1}^m$ at locations $\{\mathsf{x}_i\}_{i=1}^m$ (as shown in Fig.~\ref{fig:selfLocalizationMatching}), with cost rate $c_{ij}$ and capacity $u_{ij}$ defined using~\eqref{eq:costVisualMatch}--\eqref{eq:capacityVisualMatch}. Given navigation radius $r$ and source and target vertices $\{s,t\}$, the flows $\{y_{ij}\}$ required for self localization can be obtained by solving the following flow problem.
\begin{equation}
\begin{aligned}
& \underset{y_{ij}}{\textnormal{min}}
& &  \sum_{e_{ij}\in\mathcal{E}} c_{ij}y_{ij}, \\
&&& 0\leq {y}_{ij}\leq u_{ij},\,\,\,\,\,\,\,\,\,\,\,\,\forall e_{ij}\in\mathcal{E},\\
&&& y_{s} = q, y_{t} = -q, y_{i} = 0,\,\,\,\,\,\,\,\,\forall v_i\in \mathcal{V}\setminus (s\cup t),\\
&&& \norm{\sum_{v_{i}\in\mathcal{V}'}{\mathsf{x}_iy_{i(l+1)}} - \sum_{v_{i}\in \mathcal{V}'}{\mathsf{x}_i}y_{il}}\leq r, \forall p_l,p_{l+1}\in\mathcal{P}.
\end{aligned}
\label{eq:selfLocalization_final}
\end{equation}
\end{result}
The flow problem of~\eqref{eq:selfLocalization_final} is convex and can be solved optimally using
Second Order Cone Programming (SOCP). We use the solution of SOCP in~\eqref{eq:flowSumGeomMatch} to obtain the location $\mathsf{x}_l$ of query image $\mathcal{I}_l$ represented by vertex $p_l\in\mathcal{P}$. The proposed localization method is summarized in  Algorithm~\ref{al:selfLocalization}.

\begin{algorithm}
{\small
\caption{\small $\mathcal{L}$ = selfLocalization($\mathcal{V}'$, $\mathcal{I}_p$)}
\label{al:selfLocalization}
\begin{algorithmic}
 \STATE 1. Construct $\mathcal{G}= \{\mathcal{V},\mathcal{E}\}$ using $\mathcal{V}'$ and $\mathcal{I}_p$ (ref. Fig.~\ref{fig:selfLocalizationMatching}).
 \STATE 2. Compute rates $\{c_{ij}\}$  and capacities $\{u_{ij}\}$, using~\eqref{eq:costVisualMatch}--\eqref{eq:capacityVisualMatch}.
 \STATE 3.  Obtain flows $\{y_{ij}\}$ by solving the flow problem~\eqref{eq:selfLocalization_final}. 
 \STATE 4. Compute the location $\mathsf{x}_l$ using~\eqref{eq:flowSumGeomMatch}. Return, $\mathcal{L}=\{ \mathsf{x}_l\}_{l=1}^q$.
\end{algorithmic}
}
\end{algorithm}

\label{subsec:LandmarkExperiments}
\begin{figure*}[ht]
\centering
\includegraphics[width=\textwidth]{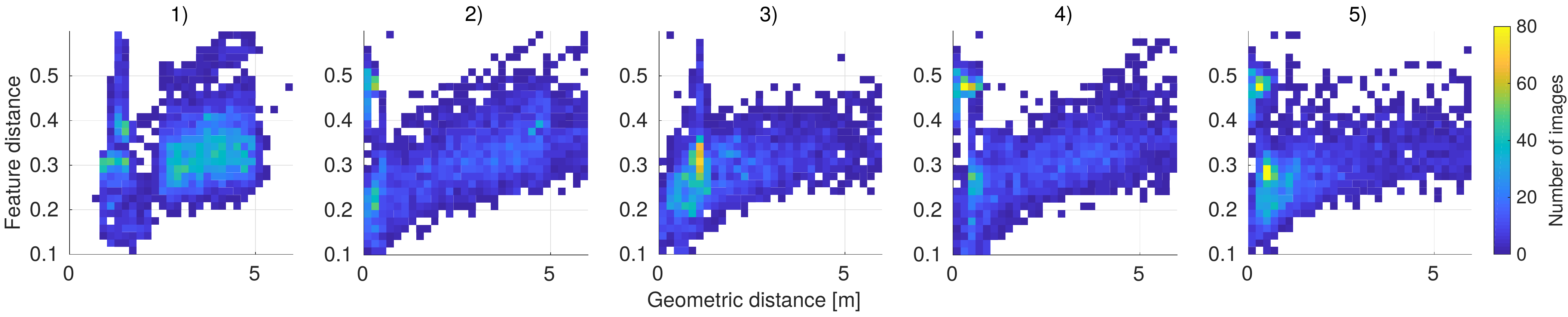}
\caption{Distribution of normalized feature distance and geometric distance of points in the full reference set $\mathcal{V}$ to the geometric nearest neighbour in the summarized reference set $\mathcal{V'}$, for the following settings from left to right: 1) Reference images sampled uniformly along sequence path. 2) Network flow, without imposing geometric representation (anchors) and visual representation (sensitivity). 3) Network flow without sensitivity. 4) Network flow without anchors. 5) Network flow imposing geometric and visual representation (our method).}
\label{fig:LandmarkSelection}
\end{figure*}

\section{Experiments}
We conduct experiments on two publicly available real world datasets, COLD-Freiburg \cite{pronobis2009ijrr} and the Oxford Robotcar \cite{maddern20171} database. The following paragraphs describe how we obtain the location coordinates $\mathcal{X}$, visual features $\mathcal{F}$ and the edges $\mathcal{E}$, as introduced in Section~\ref{subSec:problemSetup}.

\noindent\textbf{Location Coordinates.}
The COLD-Freiburg sequences directly provide location coordinates $\mathcal{X}$.
For the Oxford Robotcar dataset we use UTM coordinates, i.e.\ northing and easting.
We exclude any sequences with inaccurate or incomplete GPS and INS trajectories.
Given the large size of the Oxford Robotcar dataset and the limitation in download speed for public users, we limit ourselves to a randomly selected subset of sequences and only look at roughly the first 1250m of each run. 

\noindent\textbf{Visual Features.}
We use two different types of image features $\mathcal{F} = \{\mathsf{f}_i\}_{i=1}^n$.
The first type are VGG16~\cite{Simonyan2014} based off-the-shelf NetVLAD~\cite{arandjelovic2016netvlad} features with PCA and whitening, where the VGG16 layers were pretrained on ImageNet \cite{deng2009imagenet} and the NetVLAD weights are computed using 30'000 images from Pittsburgh~250k \cite{Torii-PAMI2015}. 
The second type of features are simply the output of the last VGG16 fully connected layer using the weights from \cite{Simonyan2014}.
The resulting feature vectors have length 4096 for NetVLAD and 1000 for VGG16 FC3.

\noindent\textbf{Edges.} For the COLD-Freiburg dataset, we look at any connection between images that are less than 2m apart. If the connection does not intersect with any walls on the given floor plan, we add it to $\mathcal{E}$. 
For the Oxford Robotcar dataset, we add a connection between two images to $\mathcal{E}$ if the distance between the images is smaller than a threshold of 12m. To avoid edges that cut corners, we use the geodesic distance.

\begin{figure*}[ht]
    \centering
    \includegraphics[width=\linewidth]{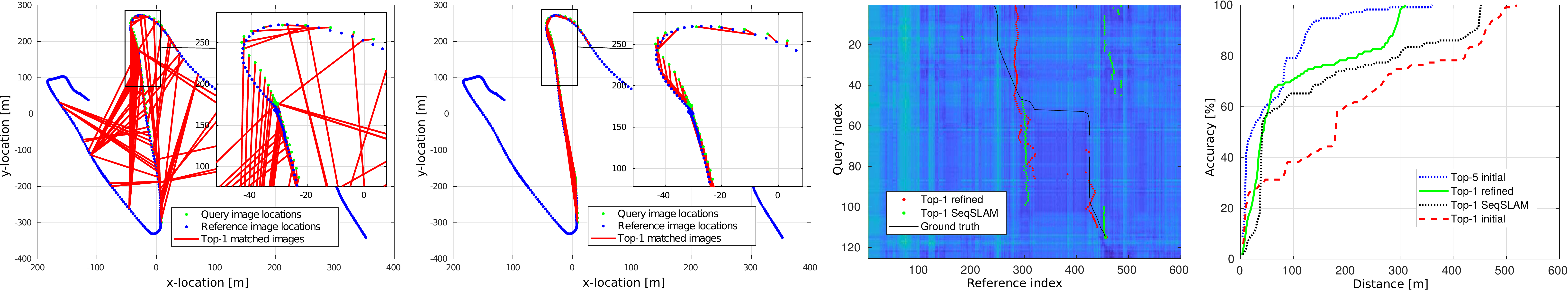}
    \caption{Left to right: Initial top-1 matches between a rainy uniformly sampled reference and an overcast query sequence of the Oxford Robotcar dataset. Top-1 matches refined by applying our self localization algorithm. Matches obtained with our method and with SeqSLAM shown on the visual distance matrix. Localization accuracy for a given distance threshold for top-1 \& top-5 localization without sequence information, SeqSLAM and our localization method.}
    \label{fig:SelfLocalization}
\end{figure*}

\subsection{Landmark Selection}
\label{subseq:LandmarkSelection}
We validate our map construction approach, introduced in Section~\ref{sec:MapConstruction}, on real world data. 
Starting from 4853 images from the first 1.25km of a rainy Oxford Robotcar sequence (2015-10-29 12:18:17), we build a reference summary $\mathcal{V'}$ with $|\mathcal{V'}| = 250$. For a total of five different setups, Fig.~\ref{fig:LandmarkSelection} shows the distribution of feature distance $\mathbf{d}(\mathsf{f},v_{k_x}')$ and geometric distance $\mathbf{d}(\mathsf{x},v_{k_x}')$ of the points in the original set $\mathcal{V}$ to the geometrically nearest neighbour $v_{k_x}'$~\eqref{eq:kx} in the summarized set $\mathcal{V'}$. First we study the baseline setup obtained by sampling uniformly along the path of the captured reference sequence.
Then we analyze our approach. 
To illustrate the impact of anchors (Section~\ref{subseq:Anchors}) and sensitivity (Section~\ref{subseq:Sensitivity}) on geometric and visual representation, we selectively switch off these two constraints.

The results in Fig.~\ref{fig:LandmarkSelection} clearly show that a reference summarized using network flow has better geometric and visual representation than a uniformly sampled baseline representation with the same number of images. We observe that introducing anchors reduces the number of points with high geometric distance and introducing sensitivity reduces the number of points with high feature distance. 

\subsection{Self-Localization}
In this section, we illustrate the feasibility of our self localization algorithm. 
As a reference set, we take 600 images from the same rainy Oxford Robotcar sequence (2015-10-29 12:18:17) as in Section~\ref{subseq:LandmarkSelection}.
We use uniform sampling as the baseline for summarizing the reference set.
As a query sequence, we uniformly sample 125 images from an overcast Oxford Robotcar sequence (2015-02-13 09:16:26) using a step size of 20 images. 
The leftmost subplot in Fig.~\ref{fig:SelfLocalization} illustrates the unrefined top-1 feature matches between the query sequence images and the reference images.
The second subplot in Fig.~\ref{fig:SelfLocalization} shows the top-1 feature matches after applying our self localization algorithm.
It is evident that our approach greatly improves the localization for this example by removing inconsistent matches. 
The third subplot in Fig.~\ref{fig:SelfLocalization} shows the visual distance matrix between query sequence and reference set (ordered according to topology, i.e.\ the originally driven route). It shows that matches do not occur independently and that the neighbours of matching images also have low feature distance. 
The true matches (i.e.\ the matches with smallest geometric distance) are indicated in black. 
In red, we plot the refined matches of our self-localization algorithm. 
As a comparison, in green we show the matches produced by SeqSLAM~\cite{milford2012seqslam}, the state of the art for capitalizing on sequential information to improve image matches.
Finally, the rightmost subplot in Fig.~\ref{fig:SelfLocalization} indicates the percentage of correctly localized images for any given distance threshold.
As an example, for a tolerance of 80m, our method has an accuracy of 68.7\%, while SeqSLAM reaches 60.9\%.

\subsection{Quantitative Evaluation}
\label{subseq:quantitative}
\begin{figure*}[ht]
\centering
\includegraphics[trim={0cm 0cm 0cm 0cm},clip,width=\textwidth]{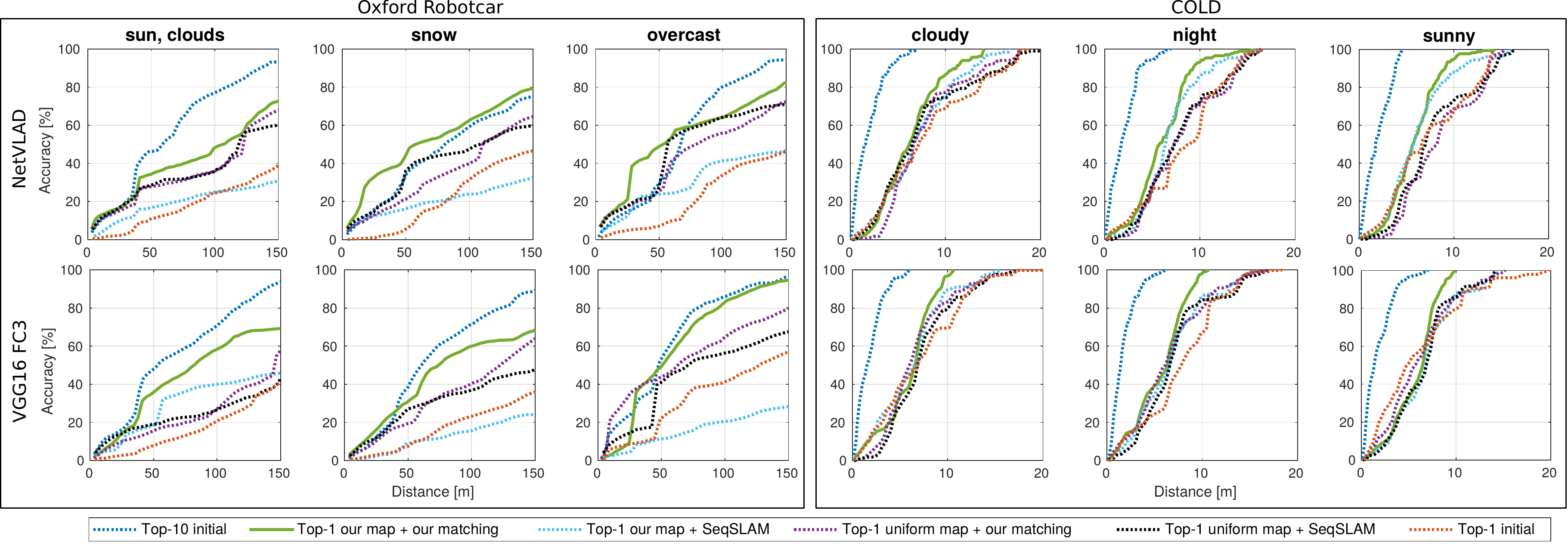}
\caption{Accuracy vs.\ distance plot for three partial sequences from the Oxford Robotcar dataset and three full sequences from the COLD-Freiburg dataset. Red and dark blue: Unrefined top-1 and top-10 matches on a uniformly summarized reference set. Black: SeqSLAM on a uniformly summarized reference set. Purple: Our self-localization on a uniformly summarized reference set. Light blue: SeqSLAM on our network flow based map. Green: Our method, with network flow based map construction and self localization.}
\label{fig:QuantitativePlot}
\end{figure*}

We provide quantitative evaluation on the COLD-Freiburg and the Oxford Robotcar dataset.
For both datasets we randomly choose one reference and three query sequences. 
For the Oxford Robotcar dataset, the reference is the rainy sequence from 2015-10-29 12:18:17.
The query sequences are taken in three different conditions: Sun and clouds (2014-11-18 13:20:12), snow (2015-02-03 08:45:10) and overcast (2015-02-13 09:16:26).
From the COLD-Freiburg dataset we use the second sunny sequence of the extended part A as a reference. As query sequences, we use the first sunny, cloudy and night sequences taken on the extended part A.

Fig.~\ref{fig:QuantitativePlot} plots the percentage of correctly localized query images for a given distance threshold for each of the six different query sequences.
The number of images in the reference set are 415 for Oxford Robotcar and 50 for COLD-Freiburg.
The results in~\ref{fig:QuantitativePlot} show that, by incorporating sequential information, SeqSLAM clearly outperforms the unrefined top-1 localization on a uniformly summarized reference set.
However, the top-1 accuracy achieved by our map building algorithm in combination with our self localization is even higher.
For some distance thresholds, the top-1 accuracy of our method even beats the unrefined top-10 reference.
The benefit of our method is especially pronounced for the more challenging Oxford Robotcar dataset.

\begin{figure}[htb]
\centering
\includegraphics[width=\linewidth]{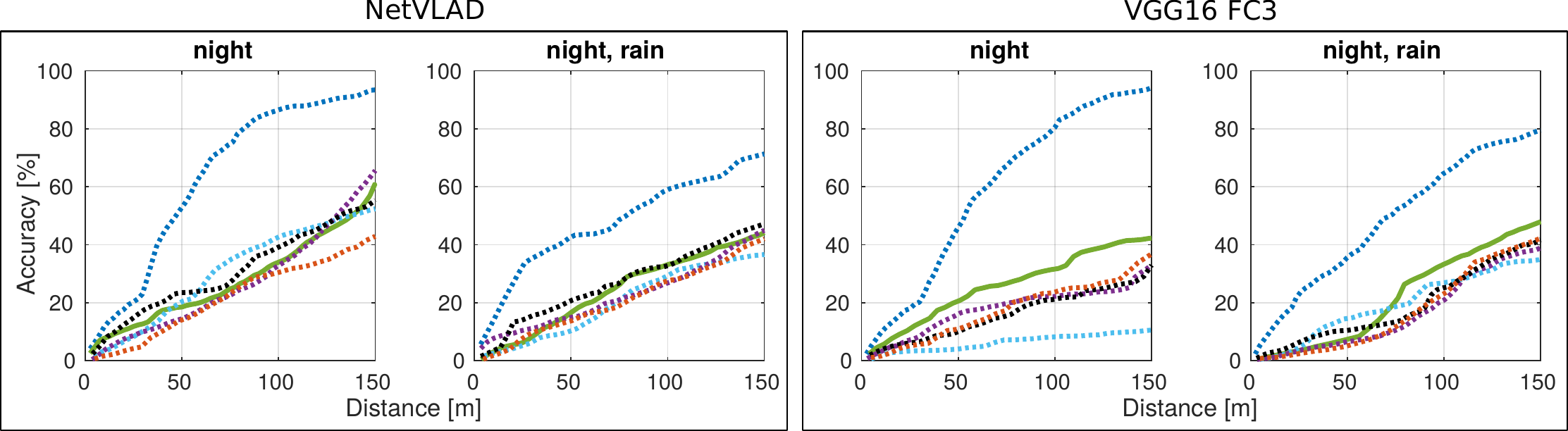}
\caption{Failure cases on Oxford Robotcar night (2014-12-16 18:44:24) and night, rain (2014-12-17 18:18:43) sequences. For legend and description see Fig.~\ref{fig:QuantitativePlot}.}
\label{fig:Fail}
\end{figure}

\begin{figure}[htb]
\centering
\includegraphics[width=\linewidth]{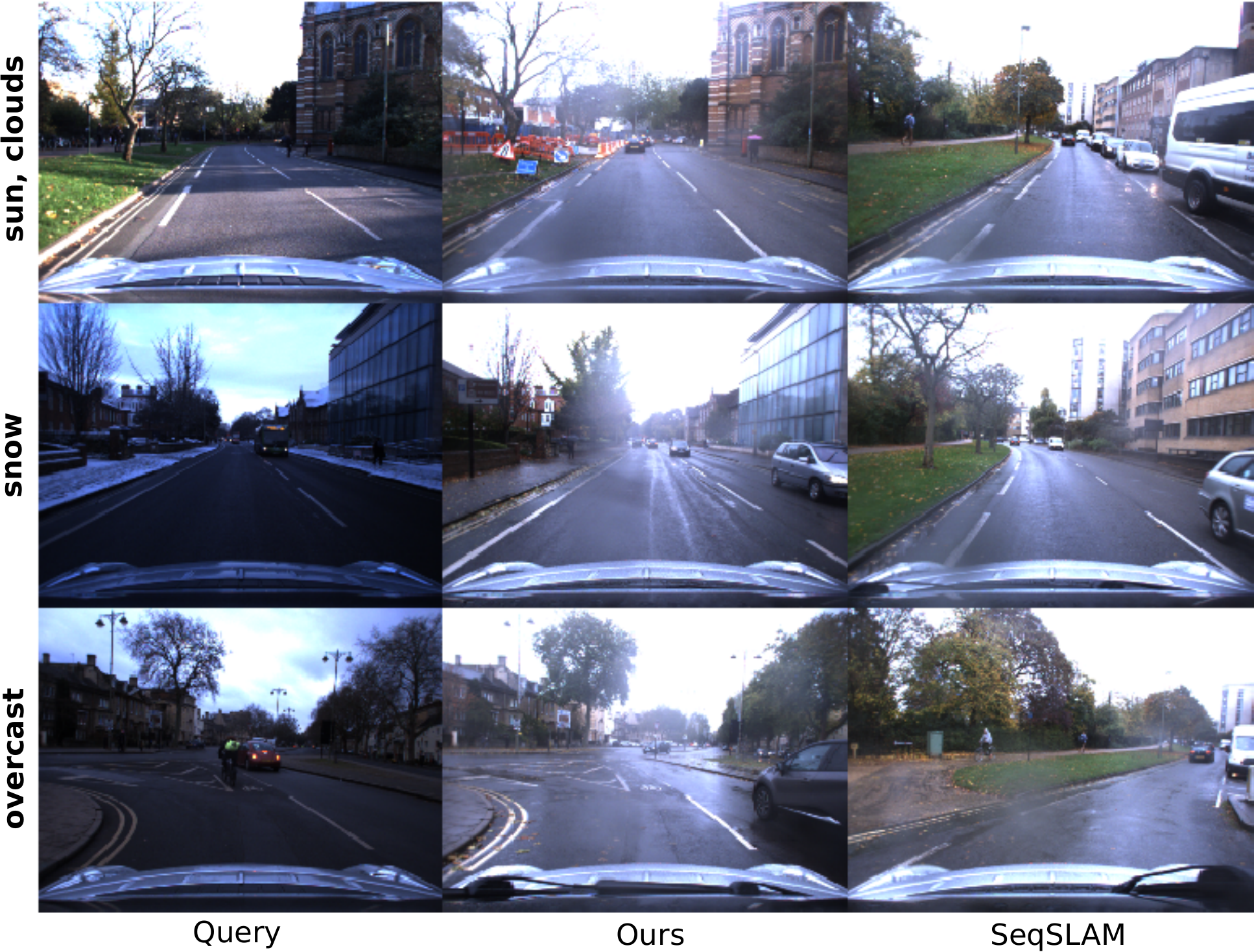}
\caption{Visual examples of our method correctly localizing, while SeqSLAM baseline fails on Oxford Robotcar dataset.}
\label{fig:Quali}
\end{figure}

While our method shows significant improvement on all sequences presented in Fig.~\ref{fig:QuantitativePlot}, it fails on sequences with non-distinctive image features, such as the outdoor night sequences in the Oxford Robotcar dataset. This is shown in Fig.~\ref{fig:Fail}. It can be observed, that for these sequences, the baseline method using SeqSLAM also fails.
Fig.~\ref{fig:Quali} shows three examples of query images which are correctly localized by our method while SeqSLAM fails.
The reference sets and query sequences are the same as the ones used for Fig.~\ref{fig:QuantitativePlot}.
The images in Fig.~\ref{fig:Quali} were matched using NetVLAD features.

\section{Conclusion}
In this paper we have formulated a set of requirements for map building and self localization in the context of image-based navigation. Based on these requirements, we proposed a method to perform map building by selecting the most suitable images for navigation. To improve self localization we proposed a method that can use multiple query images. We modeled both the methods using network flow and solved them using convex quadratic and second-order cone programs, respectively.
Our experiments on challenging real world datasets show that our approach significantly outperforms existing methods.

\paragraph{Acknowledgements.}
This research was funded by the EU's Horizon 2020 programme under grant No. 687757 -- REPLICATE  and by the Swiss Commission for Technology and Innovation (CTI), Grant No. 26253.1 PFES-ES -- \mbox{EXASOLVED}.

{\small
\bibliographystyle{ieee}
\bibliography{ms}
}
\end{document}